# Multi-Scale Deep Compressive Sensing Network

Thuong Nguyen Canh and Byeungwoo Jeon
*Department of Electronics and Computer Engineering, Sungkyunkwan University, Korea*
{ngcthuong, bjeon}@skku.edu

*Abstract*—With joint learning of sampling and recovery, the deep learning-based compressive sensing (DCS) has shown significant improvement in performance and running time reduction. Its reconstructed image, however, losses high-frequency content especially at low subrates. This happens similarly in the multi-scale sampling scheme which also samples more low-frequency components. In this paper, we propose a multi-scale DCS convolutional neural network (MS-DCSNet) in which we convert image signal using multiple scale-based wavelet transform, then capture it through convolution block by block across scales. The initial reconstructed image is directly recovered from multi-scale measurements. Multi-scale wavelet convolution is utilized to enhance the final reconstruction quality. The network is able to learn both multi-scale sampling and multi-scale reconstruction, thus results in better reconstruction quality.

## I. INTRODUCTION

Compressive sensing (CS), an emerging sampling method, facilitates a low-complexity encoder by simultaneous sampling and compression via linear projection [1]. It captures a sparse or compressible signal, $x \in N$, into a compressed form, $y \in \mathbb{R}^M, M \ll N$, via a linear projection as:

$$y = \Phi x, \qquad (1)$$

where $\Phi \in \mathbb{R}^{m \times n}$ is a sampling matrix. The Gaussian random matrix is widely used due to its theoretical guarantee but at significant computation and storage cost. In the past decades, a number of researchers have sought to alleviate the computational complexity. For example, the block-based CS (BCS) samples signals in a block-based manner [2] and Kronecker CS [3] samples each signal dimension separately.

Prior information on a signal helps a lot in its sampling or compression. Typical compression standards such as JPEG, HEVC/H.265 achieve substantially high coding efficiency by successfully exploiting prior knowledge on the to-be-compressed signals [4]. Such availability cannot be assumed easily in signal sampling stage. For example, CS cannot assume prior information on the to-be-sampled signal beyond general sparsity prior. Therefore, based on the very general observation of the human visual system, a low-frequency prior is typically assumed, which has motivated researchers to develop multi-scale CS [5-8] which captures more low-frequency components in image/video signals – input image signal is linearly decomposed into multi-scales and sampling is done adaptively to each scale.

This work is supported in part by the National Research Foundation of Korea (NRF) grant (2017R1A2B2006518) funded by the Ministry of Science and ICT.

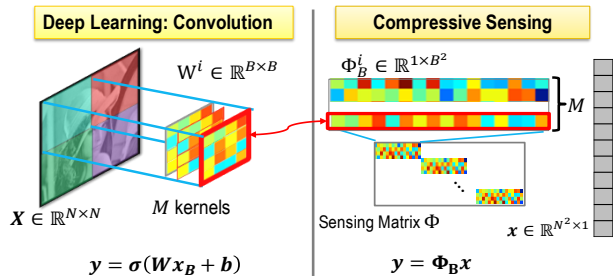

Fig. 1 Comparison between BCS and convolution layer [11]. Output of convolution layer is equivalent to BCS when $\Phi_B^i = vect(W)$, no bias (b = 0), and identity activation $\sigma$. $\Phi_B$ is a block measurement matrix.

Recently, deep learning (DL) techniques are seen to provide state-of-the-art performance in image restoration [9, 10]. Unlikely the complicated training phase of DL, its test phase is rather simple and fast. Since the linear projection of CS represented by $y = \Phi x$ can be understood as a fully connected layer having an identity activation and without bias [11], the fully connected layer shares the same practical problem of high signal dimensionality as the conventional frame-based compressed sensing. As a result, most research on DL-based CS (DCS) has focused on block-based schemes [11-14] and relied mostly on the learned prior from big data. Recently, multi-scale prior has been applied in image reconstruction as can be seen in the layer-wise wavelet network in [15], or multi-wavelet convolutional network (MWCNN) [10] which greatly improves the restoration performance.

Researches have focused on a BCS [11-14] and structured sampling with Kronecker network [20]. However, most of the previous works were about a single scale sensing. While multi-scale CS has shown significant performance gain over conventional CS though much attention [5-8], it still calls for further research on DCS for multi-scale sampling. All of these observations motivated this paper to exploit DL-based method for multi-scale sampling scheme. Firstly, we linearly decompose images into multi-scales by wavelet decomposition. Secondly, we sample the signals across all scales and reconstruct an initial image. Lastly, we enhance the reconstruction performance with MWCNN [10]. By jointly learning the multi-scale sampling matrix and multi-scale reconstruction, we can produce better quality than the state-of-the-art conventional (single and multi- scales) and DL schemes.

This paper is organized as follows. We review related works in multi-scale sampling and deep learning CS in Section 2. Section 3 proposes our multi-scale DCS network (MS-DCSNet) with multiple phases of training. We evaluate our proposed method in Section 4 and draw some conclusion in Section 5.

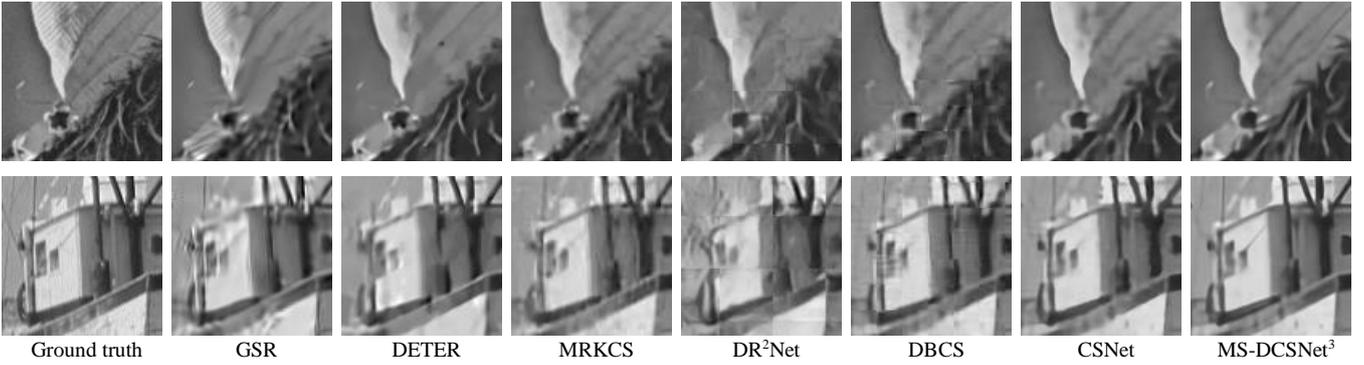

Fig. 2 Cropped reconstructed Lena and Boats images (size 512x512 at subrate 0.1) in the first and second rows, respectively. Aliasing artifact is observed at MRKCS, DBCS, and CSNet. MS-DCSNet[3] (see IV.A) shows the best visual quality of reconstruction (but still shows loss of details around Lena's hat).

## II. RELATED WORK

### A. Multi-Scale Compressive Sensing

Recent works [5-8] have proven that multi-scale CS can be the optimal sampling solution. Radial Fourier subsampling [1] can be considered as a simple example and often used in bio-imaging due to its physically driven projection. Another example is the block-based multi-scale sampling in MS-SPL [6] which first transforms image into wavelet domain, and then uses adaptive sample rate for each decomposition level. Canh et al. developed multi-scale sampling for Kronecker CS in [7, 8]. The authors first decomposed images into various scales by wavelet, pyramid, or multiple-resolution decomposition, and adaptively sampled each scale. The decomposition which is linear can easily incorporate sampling matrix. In addition, low-frequency components are captured more often. In DCS, we ensure all sampling-related layers to be linear by using identity activation and without bias for convolution.

### B. Deep Compressive Sensing

DCS [11] attempts to recover images from compressed measurements. Due to complexity of frame-based sampling, more researchers followed the BCS scheme which learns to recover a small block of signals [11, 12] with a fixed sensing matrix Φ. Authors [11, 12] showed that DCS can reconstruct image but with a limited performance. However, it is possible to jointly learn the sampling matrix together with reconstruction [13]. Note that, early methods [11-13] use block-wise sampling and reconstruction, thus resulting in blocking artifacts in the reconstructed image. By modeling BCS as convolution layer (as in Fig. 1), CSNet [14] enabled block-based sampling with frame-based reconstruction. It, therefore, achieved better reconstruction performance without the blocking artifacts.

Since the sampling matrix is also learned [13, 14], the trained sampling matrix tends to capture more low-frequency information similarly to the multi-scale sampling scheme. This is shown to cause loss of high-frequency content and/or aliasing artifact in the reconstructed images as visualized in Fig. 2. The aliasing artifact is clearly visible in Boat image reconstructed by conventional (MRKCS) and DL (CSNet, DBCS) methods.

### C. Multi-Level Wavelet Convolution

Recently, wavelet decomposition is employed in deep learning as an efficient domain for image restoration [10, 15]. Conventionally, a 2D input is decomposed into 4 sublayers and convolution is applied independently for each layer sequentially [15]. Researchers [10], further developed the Multi-level Wavelet convolutional (MWCNN) which applies convolutional neural network (CNN) across all wavelet layers. Despite the orthogonality of each sub-band, structural similarity still exists among layers. Therefore, CNN is able to extract features across multiple frequencies. We utilize this for multi-scale sampling and multi-scale reconstruction.

## III. PROPOSED MULTI-SCALE DEEP COMPRESSIVE SENSING

In this section, we propose a multi-scale deep compressive sensing network, named as MC-DCSNet[*], shown in Fig. 3.

### A. Multi-Scale Wavelet Sampling

To perform multi-scale sampling, a Discrete Wavelet Transform (DWT) layer is generated. With a given image of size $n \times n$, DWT outputs wavelet coefficients at four frequency bands in a form of $\frac{n}{2} \times \frac{n}{2} \times 4$. Conventional multi-scale sampling scheme samples each scale independently at a proper rate like [6, 7]. We use the MWCNN concept (i.e., utilize correlation among frequency bands) for multi-scale CS by sampling measurements across all wavelet scales under BCS scheme. BCS is implemented as $m_i$ convolutions with kernel $n_B \times n_B \times 4$, no bias and no activation [14]. $n_B$ denotes the block size. Each measurement contains information about all frequency bands and thus helps learning to sample multi-scale.

### B. Initial Reconstruction

Similar to CSNet [14], we recover block measurement by $1 \times 1$ convolution followed by a reshape and concatenate layer. Instead of recovering each wavelet scale separately and then using inverse discrete wavelet transform (iDWT) to obtain initial reconstructed image [7], we directly recover image at the original size by $4B^2$ convolution 1x1 as shown in Fig. 3.

---

[*]The source code for our MS-DCSNet and test image sets are available at github.com/AtenaKid/MS-DCSNet-Release.

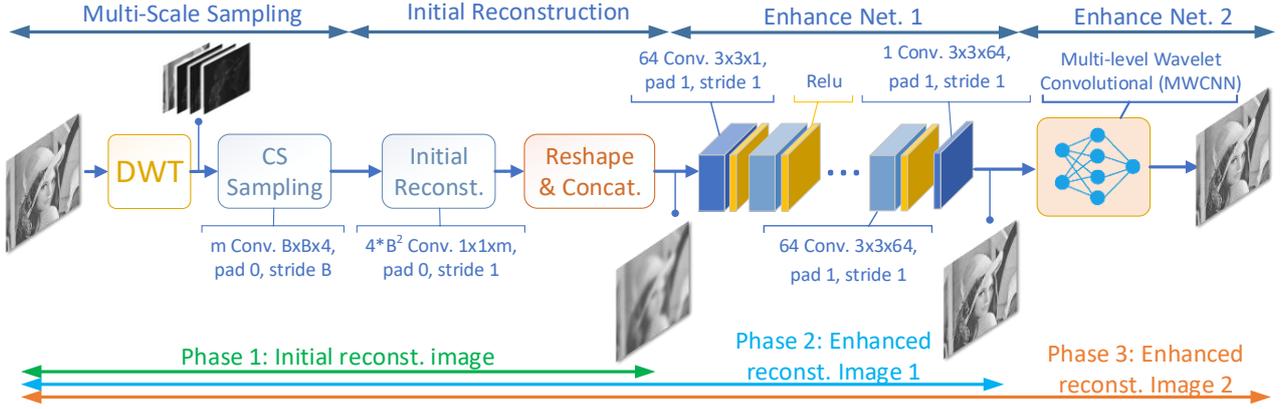

Fig. 3 Proposed Multi-scale Deep Compressive Sensing Network (MS-DCSNet). MWCNN [10] is used as the 2nd enhanced network.

## C. Multi-Scale Reconstruction (Enhance Network)

After obtaining the initial reconstructed image, enhanced reconstruction network is used. Our enhance network consists of two sub networks. The first enhance network is a simple convolution network with a series of convolutions (3x3) and ReLu as depicted in Fig. 3. The network structure is similar to CSNet so that we can make fair comparison. To further improve the reconstruction performance, the second enhance network is used. In this work, we adopt the multi-level wavelet decomposition with convolution combination in MWCNN [10] which has shown superior performance in image restoration including image denoising and super-resolution. In this paper, our proposed MS-DCSNet only uses few layers structure such as convolution, rectified linear unit, and concatenation.

## D. Training Network

**Loss function**. As the proposed MS-DCSNet follows end-to-end learning structure, inputs and labels are identical as the ground truth image. Similar to many image restoration methods [8-14], Euclidean loss is used as an objective function as:

$$\min \frac{1}{2N} \sum_{i=1}^{N} \|f(x_i, \theta) - x_i\|_2^2, \quad (3)$$

where $N$ denotes the total number of training samples, $x_i$ represents a sample, and $f$ is the network function at a setting $\theta$.

**Multi-phases training**. As shown in [12], the better initial image reconstruction is, the better final reconstruction quality can be. Authors utilized two phases training which first learns initial reconstruction then enhances reconstruction. This paper proposes a training process of three-phases of (Phase 1) initial reconstruction, and two enhanced reconstruction phases (Phase 2, 3). In each phase, we use learning rate 0.001, 0.0001 and 0.00005 for each 30 epochs. It should be noted that, only the Phase 3 uses the multi-scale prior following the MWCNN structure [10]. For Phase 3 with MWCNN, we used the pre-trained network of Gaussian denoising at a noise level 15 as the initialization. It should be noted that, the weight in the Phase $i$ is also re-learned in the next Phase $i + 1$. In training process, we use the adaptive moment estimation (adam) as an optimization method.

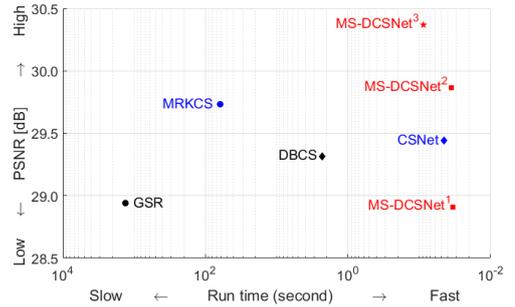

Fig. 4. Running time vs. average PSNR of classic512 at subrate 0.1.

## IV. EXPERIMENTAL RESULTS

### A. Simulation Setting

This work uses DIV2K [16] for training with 64x500 patches of size 256x256 and implement with MatConvNet [18], tested with 6 test images of classic512, Set5, and Set14[(*)]. For conventional CS, we use BCS with GSR [18], Kronecker CS with DETER [19], and multi-scale KCS with MRKCS [7]. For DL-based CS, we use single scale BCS as ReconNet [11], DR²Net [12], DBCS [13], and CSNet [14]. The tested block sizes are 33×33 (ReconNet, DR²Net), 16×16 (DBCS), 32×32 (CSNet), and 16×16×4 (MS-DCSNet) for fair comparison.

### B. Multiple-Phases Training

We separately name our network after each phase as MS-DCSNet[1], MS-DCSNet[2], and MS-DCSNet[3], and show the incremental performance in Table 2. Thanks to multi-scale sampling, even with similar network structure as CSNet, MS-DCSNet[2] still offers 0.14~0.31dB gain in Set5 and Set14 [14].

### C. Performance Comparison

Tables 1 and 2 show that the joint learning recovery and sampling schemes (DBCS, CSNet, and MS-DCSNet) outperform those of learning recovery (ReconNet and DR²Net). DBCS and CSNet show 0.4~1.5 dB gain over conventional BCS but less than the frame-based (DETER) and multi-scale sampling (MRKCS). MS-DCSNet[3] performs the best with 0.62~1.37 dB gain over CSNet and MRKCS in the 512x512 test images with little sacrifice in running time as in Fig. 4.

Table 1. Performance comparison in PSNR [dB] and SSIM for various test images of size 512x512

| Image | Rate | GSR PSNR | GSR SSIM | DETER PSNR | DETER SSIM | MRKCS PSNR | MRKCS SSIM | ReconNet PSNR | ReconNet SSIM | DR²Net PSNR | DR²Net SSIM | DBCS PSNR | DBCS SSIM | CSNet PSNR | CSNet SSIM | MS-DCSNet[3] PSNR | MS-DCSNet[3] SSIM |
|---|---|---|---|---|---|---|---|---|---|---|---|---|---|---|---|---|---|
| Lena | 0.1 | 30.97 | 0.866 | 32.32 | 0.869 | 32.87 | 0.821 | 26.89 | 0.749 | 28.65 | 0.800 | 31.35 | 0.820 | 32.15 | 0.879 | **32.90** | **0.891** |
| | 0.2 | 34.44 | 0.914 | 35.31 | 0.911 | 35.85 | 0.919 | - | - | - | - | 34.34 | 0.918 | 35.21 | 0.924 | **35.86** | **0.929** |
| | 0.3 | 36.47 | 0.936 | 37.09 | 0.932 | 37.72 | 0.940 | - | - | - | - | 36.35 | 0.939 | 37.33 | 0.945 | **37.85** | **0.948** |
| Peppers | 0.1 | 31.45 | 0.843 | 32.24 | 0.844 | 33.40 | 0.853 | 26.22 | 0.716 | 28.32 | 0.769 | 32.51 | 0.863 | 32.06 | 0.858 | **33.48** | **0.870** |
| | 0.2 | 34.12 | 0.880 | 34.62 | 0.880 | **35.52** | 0.891 | - | - | - | - | 34.77 | 0.895 | 34.42 | 0.891 | 35.47 | **0.899** |
| | 0.3 | 35.65 | 0.905 | 35.99 | 0.903 | 36.59 | 0.909 | - | - | - | - | 35.97 | 0.911 | 35.84 | 0.910 | **36.68** | **0.915** |
| Mandrill | 0.1 | 19.93 | 0.508 | 20.16 | 0.449 | 21.92 | 0.549 | 19.70 | 0.411 | 20.18 | 0.455 | 22.15 | 0.587 | 22.26 | 0.592 | **22.50** | **0.610** |
| | 0.2 | 22.22 | 0.682 | 22.28 | 0.607 | 23.61 | 0.688 | - | - | - | - | 24.05 | 0.750 | 24.08 | 0.749 | **24.44** | **0.767** |
| | 0.3 | 23.92 | 0.775 | 24.06 | 0.714 | 25.13 | 0.780 | - | - | - | - | 25.69 | 0.831 | 25.72 | 0.833 | **26.05** | **0.842** |
| Boats | 0.1 | 27.55 | 0.773 | 27.46 | 0.741 | 28.78 | 0.786 | 24.35 | 0.636 | 25.64 | 0.688 | 28.50 | 0.801 | 29.08 | 0.812 | **29.66** | **0.833** |
| | 0.2 | 31.34 | 0.862 | 30.89 | 0.836 | 31.88 | 0.865 | - | - | - | - | 31.36 | 0.868 | 32.05 | 0.884 | **32.82** | **0.896** |
| | 0.3 | 33.72 | 0.904 | 33.06 | 0.883 | 33.73 | 0.899 | - | - | - | - | 33.22 | 0.882 | 33.98 | 0.911 | **34.73** | **0.919** |
| Camera-man | 0.1 | 32.12 | 0.913 | 33.62 | 0.862 | **34.15** | 0.928 | 26.03 | 0.798 | 28.46 | 0.848 | 32.20 | 0.920 | 31.15 | 0.918 | 33.10 | **0.942** |
| | 0.2 | 37.15 | 0.958 | 37.25 | 0.954 | 38.92 | 0.967 | - | - | - | - | 36.98 | 0.952 | 34.59 | 0.961 | **39.46** | **0.980** |
| | 0.3 | 40.58 | 0.977 | 39.78 | 0.976 | 42.37 | 0.984 | - | - | - | - | 40.54 | 0.978 | 37.47 | 0.976 | **44.10** | **0.992** |
| Man | 0.1 | 27.74 | 0.781 | 28.14 | 0.760 | 29.58 | 0.812 | 25.30 | 0.660 | 26.51 | 0.714 | 29.07 | 0.812 | 29.84 | 0.833 | **30.21** | **0.847** |
| | 0.2 | 30.63 | 0.867 | 31.20 | 0.847 | 32.31 | 0.885 | - | - | - | - | 31.89 | 0.891 | 32.55 | 0.907 | **32.87** | **0.914** |
| | 0.3 | 32.83 | 0.921 | 33.41 | 0.895 | 34.33 | 0.924 | - | - | - | - | 33.37 | 0.917 | 34.52 | 0.939 | **34.94** | **0.944** |
| Average | 0.1 | 28.29 | 0.781 | 28.99 | 0.754 | 30.11 | 0.792 | 24.75 | 0.662 | 26.29 | 0.712 | 29.30 | 0.801 | 29.42 | 0.815 | **30.31** | **0.832** |
| | 0.2 | 31.65 | 0.861 | 31.93 | 0.839 | 33.02 | 0.869 | - | - | - | - | 32.23 | 0.879 | 32.15 | 0.886 | **33.49** | **0.898** |
| | 0.3 | 33.86 | 0.903 | 33.90 | 0.884 | 34.98 | 0.906 | - | - | - | - | 34.19 | 0.910 | 34.14 | 0.919 | **35.73** | **0.927** |

Table 2. Average PSNR [dB] & SSIM values by various algorithms on Set5 and Set14

| Image Set | Rate | GSR PSNR | GSR SSIM | ReconNet PSNR | ReconNet SSIM | DR²Net PSNR | DR²Net SSIM | DBCS PSNR | DBCS SSIM | CSNet PSNR | CSNet SSIM | MS-DCSNet[1] PSNR | MS-DCSNet[1] SSIM | MS-DCSNet[2] PSNR | MS-DCSNet[2] SSIM | MS-DCSNet[3] PSNR | MS-DCSNet[3] SSIM |
|---|---|---|---|---|---|---|---|---|---|---|---|---|---|---|---|---|---|
| Set5 | 0.1 | 29.98 | 0.866 | 25.98 | 0.734 | 27.79 | 0.798 | 31.31 | 0.894 | 32.30 | 0.902 | 30.66 | 0.855 | 32.44 | 0.904 | **33.39** | **0.917** |
| | 0.2 | 34.17 | 0.926 | - | - | - | - | 34.55 | 0.939 | 35.63 | 0.945 | 34.06 | 0.924 | 35.82 | 0.947 | **36.56** | **0.951** |
| | 0.3 | 36.38 | 0.949 | - | - | - | - | 36.54 | 0.956 | 37.90 | 0.963 | 36.51 | 0.952 | 38.20 | 0.965 | **38.74** | **0.967** |
| Set14 | 0.1 | 27.51 | 0.771 | 24.18 | 0.640 | 24.38 | 0.706 | 28.54 | 0.814 | 28.91 | 0.812 | 27.81 | 0.778 | 29.10 | 0.815 | **29.67** | **0.828** |
| | 0.2 | 31.20 | 0.865 | - | - | - | - | 31.21 | 0.885 | 31.86 | 0.891 | 30.69 | 0.874 | 32.05 | 0.893 | **32.51** | **0.900** |
| | 0.3 | 33.71 | 0.907 | - | - | - | - | 33.08 | 0.926 | 33.99 | 0.928 | 32.86 | 0.917 | 34.30 | 0.930 | **34.71** | **0.934** |

Some artefact due to loss of high frequency is observed for both DBCS, CSNet, MRKCS as well as MS-DCSNet[3] in image having strong edge such as in Boats in Fig. 2. MS-DCSNet[3], however, shows visually pleasing results with less aliasing. Conventional methods (GSR and DETER) preserve structures well but suffer from artifact and over-smoothing.

## V. CONCLUSIONS

In this paper, we proposed a multi-scale deep compressive sensing network (MS-DCSNet) to improve the sampling efficiency of compressive sensing. The proposed method not only utilizes multi-scale wavelet prior at sampling but also at reconstruction stage. A multi-phase training scheme is employed to improve the training efficiency. MS-DCSNet is shown to improve performance over deep learning-based CS in both subjective and objective quality.